\documentclass{article}
\usepackage{conference,times}

\usepackage{amsmath,amsfonts,bm}

\def\eqref#1{equation~\ref{#1}}

\def\1{\bm{1}}

\DeclareMathAlphabet{\mathsfit}{\encodingdefault}{\sfdefault}{m}{sl}
\SetMathAlphabet{\mathsfit}{bold}{\encodingdefault}{\sfdefault}{bx}{n}

\usepackage{graphicx}
\usepackage{booktabs}
\usepackage{multirow}
\usepackage{makecell}
\usepackage{array}
\usepackage{adjustbox}
\usepackage{algorithm}
\usepackage{algpseudocode}
\usepackage{wrapfig}
\usepackage[table]{xcolor}
\usepackage{amssymb}

\definecolor{conferenceblue}{rgb}{0.10,0.35,0.65}
\usepackage{hyperref}
\usepackage{url}

\title{\mbox{ForeFly: A Dual-Horizon World Action Model} \\
\mbox{for Aerial Vision-Language Navigation}}

\author{
\textbf{Kunhui Wang}\textsuperscript{1,2 \thanks{Equal contribution.}} \quad
\textbf{Xintong Zhang}\textsuperscript{3 \footnotemark[1]} \quad
\textbf{Junyu Gao}\textsuperscript{1,2 \thanks{Corresponding author.}} \quad
\textbf{Changsheng Xu}\textsuperscript{1,2,4} \\
\\[3pt]
\small
\textsuperscript{1}State Key Laboratory of Multimodal Artificial Intelligence Systems, \\
Institute of Automation, Chinese Academy of Sciences, Beijing, China
\\[1pt]
\textsuperscript{2}School of Advanced Interdisciplinary Sciences, \\
University of Chinese Academy of Sciences, Beijing, China
\\[1pt]
\textsuperscript{3}Division of Natural and Applied Sciences, Duke Kunshan University, Suzhou, China
\\[1pt]
\textsuperscript{4}Peng Cheng Laboratory, ShenZhen, China
\\[3pt]
\texttt{wangkunhui23@mails.ucas.ac.cn, xz405@duke.edu}
\\
\texttt{\{junyu.gao, csxu\}@nlpr.ia.ac.cn}
}

\conferencefinalcopy

\begin{document}

\maketitle
\fancyhead{}
\renewcommand{\headrulewidth}{0pt}
\setlength{\headsep}{30pt}

\begin{abstract}

Aerial Vision-Language Navigation (AVLN) requires UAVs to maintain reliable instruction following over long trajectories in complex 3D environments. 
However, existing AVLN approaches are predominantly reactive or limited to single-horizon prediction, overlooking complementary future cues across different temporal horizons.
To address this limitation, we propose ForeFly, a dual-horizon latent world action model that predicts both a proximal future for local continuity and an adaptive route-critical future for long-range guidance.
Horizon-specific foresight queries are primed with recent and route-critical visual memories, providing history-aware context for future prediction. 
To exploit their distinct roles in action generation, we introduce Foresight-Guided Action Refinement (FGAR), which asymmetrically exploits proximal foresight for local action enhancement and route-critical foresight for feature-wise correction and route-level guidance.
Experiments on the TravelUAV and UAV-ON benchmarks show that ForeFly consistently outperforms strong baselines across seen and unseen settings, validating the effectiveness of dual-horizon foresight and FGAR learning.
The code is available at: \url{https://github.com/kunhuiW/ForeFly}

\end{abstract}

\section{Introduction}
\label{sec:intro}

Aerial Vision-Language Navigation (AVLN) aims to enable unmanned aerial vehicles (UAVs) to understand natural-language instructions and autonomously navigate in complex, open-world three-dimensional environments~\cite{aerialvln,fan2022aerialVDN}. 
Compared with conventional embodied navigation on the ground~\cite{anderson2017r2r,rxr,krantz2020beyond,reverie,blukis2018mapping}, AVLN poses challenges due to the large-scale search space, continuous 3D motion, drastic viewpoint changes, and weak geometric constraints in aerial scenes~\cite{touchdown2019,gao2026openfly,lin2025openvln,Dai2026DRL}.
Moreover, language instructions in aerial navigation involve distant landmarks, route-level spatial relations, and long-horizon events, requiring the agent to maintain coherent visual-language understanding over extended trajectories~\cite{aerialvln,sun2026autofly,lee2024citynav,Gu2026UAVbasedMO}. 
These characteristics make AVLN a challenging yet important task for building robust autonomous aerial agents.

\begin{figure}[!t]
    \centering
    \includegraphics[width=0.98\linewidth]{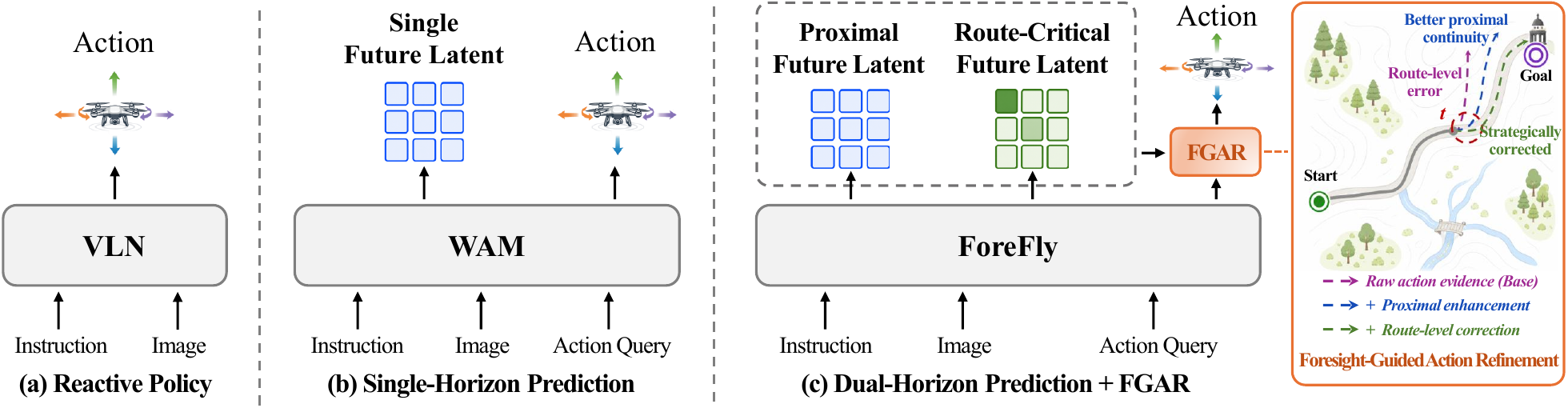}
    \vspace{-2mm}
    \caption{
    (a) Reactive policies predict actions without future modeling. 
    (b) Single-horizon methods predict a single future latent, overlooking complementary navigation roles for local continuity and route-level anticipation.
    (c) ForeFly jointly predicts proximal and route-critical futures, while FGAR integrates them for locally smooth and strategically consistent navigation.
    \vspace{-4mm}
    }
    \label{fig:motivation}
\end{figure}

Recent advances in vision-language navigation~\cite{gao2026openfly,zheng2026thinklikepilotfinegrained,lin2025openvln,jiang2026spatialfly} have explored end-to-end frameworks to align visual observations with language instructions and translate multimodal understanding into navigation actions.
Existing methods typically encode the current view, instruction context, and trajectory history to predict the next waypoint or control command, enabling UAVs to follow complex routes in outdoor environments.
Despite their progress, most of these methods predominantly operate within a reactive paradigm (Figure~\ref{fig:motivation}(a)), where actions are inferred mainly from already observed visual-language evidence.
As a result, they lack an explicit mechanism to anticipate future visual states and navigation-critical events before they become immediately observable.

The absence of anticipatory modeling is particularly problematic in aerial navigation. 
Under continuous 3D motion and rapidly changing viewpoints, local decision errors tend to propagate into substantial trajectory deviations~\cite{lin2025openvln,wang2024openuav}.
Missing a turning point or failing to identify a landmark transition can cause the agent to drift away from the instruction-relevant route, making subsequent recovery increasingly difficult.
Therefore, effective AVLN agents should go beyond reactive action prediction and reason about instruction-relevant future states.

A natural way to introduce such foresight is to predict future observations.
However, directly forecasting future images in the pixel space is computationally demanding and will force the model to reconstruct visual details that are irrelevant to navigation, such as textures, illumination, or background appearance~\cite{chen2025largevideoplanner,emami2025diffusion,zhao2026worldvln}.
Meanwhile, existing single-horizon prediction typically focuses on the immediate future, capturing short-term visual continuity but overlooking complementary cues at different temporal horizons that are critical for long-horizon aerial decision-making, such as upcoming turns, landmark transitions, and target-approaching regions~\cite{jia2026driveworldvla,zhang2026minddriver,zheng2026worldfly}.
Motivated by this observation, we model navigation-relevant futures at two complementary temporal horizons, predicting both proximal and route-critical future states in a compact latent space instead of reconstructing them in the pixel space.
Specifically, proximal futures capture local motion continuity under continuous 3D motion and drastic viewpoint changes, while route-critical futures anticipate sparse navigation events to support long-range guidance over large-scale, weakly constrained aerial environments.

To this end, we propose \textbf{ForeFly} (Figure~\ref{fig:motivation}(c)), a retrospective dual-horizon world action model for AVLN. ForeFly first grounds horizon-specific foresight in recent and route-critical visual memories, 
then predicts complementary proximal and adaptive route-critical future latents to capture short-term motion continuity and long-range route anticipation, 
finally refines waypoint generation through Foresight-Guided Action Refinement (FGAR), which assigns asymmetric decision roles to dual-horizon foresight, leveraging proximal foresight for local action enhancement and route-critical foresight for strategic route-level correction. Our contributions are threefold:
\begin{itemize}

  \item[$\bullet$] We reformulate AVLN from reactive action prediction into dual-horizon foresight that jointly models complementary proximal and route-critical latent futures, supporting both local motion continuity and long-range route guidance. 
  \item[$\bullet$] We instantiate dual-horizon foresight with horizon-specific foresight query priming and an asymmetric Foresight-Guided Action Refinement (FGAR) module, which leverages complementary historical cues and translate proximal and route-critical foresight into local action enhancement and route-level correction, respectively.
  \item[$\bullet$] Experiments on TravelUAV and UAV-ON show consistent gains over strong baselines, particularly on long-range and unseen trajectories, while ablations validate the effectiveness and complementarity of dual-horizon foresight.

\end{itemize}

\section{Related Work}
\label{sec:Related Work}

\noindent{\bf Aerial Vision-Language Navigation.} 
Vision-Language Navigation (VLN) was initially studied in ground-based environments, 
where embodied agents follow natural-language instructions through indoor or street-level scenes~\cite{anderson2017r2r, rxr, ETPNav, Zhu2020VLN_auxiliary, Wang2021SSM}.
Compared to ground-based VLN, AVLN presents unique challenges, 
including a continuous 3D motion space, drastic viewpoint transitions, and heavy reliance on long-range landmark dependencies~\cite{xu2026aerialvln,chen2026visionandlanguagenavigationuavsprogress}.
To address these challenges, early AVLN methods improve visual-language grounding through auxiliary objectives, such as spatial-relation matching and contrastive learning~\cite{su2025fela,GeoText1652}. 
More recent approaches increasingly adopt Multimodal Large Language Models (MLLMs) for navigation reasoning.
For example, TravelUAV~\cite{wang2024openuav} and FlightVLA~\cite{zheng2026thinklikepilotfinegrained} employ hierarchical architectures that combine high-level language planning with low-level action refinement.
Other methods incorporate depth or implicit 3D cues to strengthen spatial reasoning~\cite{sun2026autofly,Zhu2026UAV3S,jiang2026spatialfly,Sun2026GroundToSpace,Liu2026KnowingTheSelf}, while several studies exploit historical spatial-semantic evidence to improve temporal consistency over long trajectories~\cite{kim2025raven,lou2026voln,qi2026parse,ding2026history}.
Despite this progress, most existing AVLN methods remain predominantly reactive, predicting immediate actions from current and historical observations without explicitly modeling future world states. Consequently, they lack an explicit mechanism to anticipate navigation-critical events and use such foresight for proactive decision-making.

\noindent{\bf World Action Model for VLN.} 
World models have recently emerged as a promising paradigm for VLN, learning environment dynamics from past observations and actions to predict future states for navigation~\cite{zhang2026minddriver,koh2021pathdreamer,NWM,Yao2025NavMorph}. 
Building on this idea, World Action Models (WAMs) unify future-state modeling and action generation within a single framework, and have recently been extended to VLN~\cite{ye2026worldactionmodelszeroshot,cen2025worldvla,xu2026wamflow}. 
Some approaches jointly optimize future-state prediction and action generation, explicitly coupling navigation decisions with their anticipated perceptual consequences~\cite{Yang2026WAMNav,2026futureNav}. 
Others employ video diffusion models to generate instruction-conditioned future observations and transfer their spatiotemporal priors to action or trajectory prediction~\cite{Liu2026ImagineUAV,xu2025aeroact}. 
Recent methods further incorporate action-aware reinforcement learning to align navigation policies with long-term rollout outcomes~\cite{zhang2026anticipateacting,zhao2026worldvln}.
Despite this progress, existing methods largely emphasize short-horizon or single-step prediction, with limited mechanisms for anticipating sparse but navigation-critical future events. 
As a result, they may fail to foresee upcoming turns, landmark transitions, or other route-level cues, limiting proactive decision-making over long trajectories.

\section{Method}

\subsection{Problem Formulation}
We formalize aerial vision-language navigation as a language-guided sequential decision-making problem, where a UAV navigates in a continuous 3D environment according to a natural-language instruction.
Given a language instruction $\mathcal{I}$, the UAV maintains a trajectory history $\mathcal{P}_{\leq t}=\{{p}_1, {p}_2,\ldots,{p}_t\}$ and
a sequence of historical observations $\mathcal{O}_{\leq t}=\{o_1,o_2,\ldots,o_t\}$.
At each time step $t$, given the navigation context
$\mathcal{C}_t=(\mathcal{I},\mathcal{P}_{\leq t},\mathcal{O}_{\leq t})$, the proposed world action model predicts dual-horizon future representations and conditions the next waypoint action on this foresight:
\begin{equation}
p_{\theta}
\left(
Z^{p}_{t+1},
Z^{c}_{t+\kappa_t},
a_t
\mid C_t
\right)
=
p_{\theta}^{W}
\left(
Z^{p}_{t+1},
Z^{c}_{t+\kappa_t}
\mid C_t
\right)
p_{\theta}^{A}
\left(
a_t
\mid
C_t,
Z^{p}_{t+1},
Z^{c}_{t+\kappa_t}
\right),
\label{eq:wam_factorization}
\end{equation}
where $\mathbf{Z}_{t+1}^{p}\in\mathbb{R}^{N_p\times d}$ denotes the proximal future latent
and $\mathbf{Z}_{t+\kappa_t}^{c}\in\mathbb{R}^{N_c\times d}$ denotes the route-critical future latent.
Here, $\kappa_t$ is not a fixed constant, but an adaptive temporal offset determined by
the route-critical target construction described in Sec.~\ref{sec:anchor_target}.
The waypoint action is represented in the current UAV coordinate frame as: 
\begin{equation}
\mathbf{a}_t=
(\mathbf{d}_t,\Delta \rho_t)
=(\Delta x_t,\Delta y_t,\Delta z_t,\Delta \rho_t)
\in\mathbb{R}^{4}.
\end{equation}
where $\mathbf{d}_t=(\Delta x_t,\Delta y_t,\Delta z_t)$ denotes the normalized relative waypoint direction in 3D space with $\|\mathbf{d}_t\|_2=1$, and $\Delta \rho_t\geq 0$ denotes the movement magnitude.

\begin{figure}[t]
  \centering
  \includegraphics[width=\textwidth]{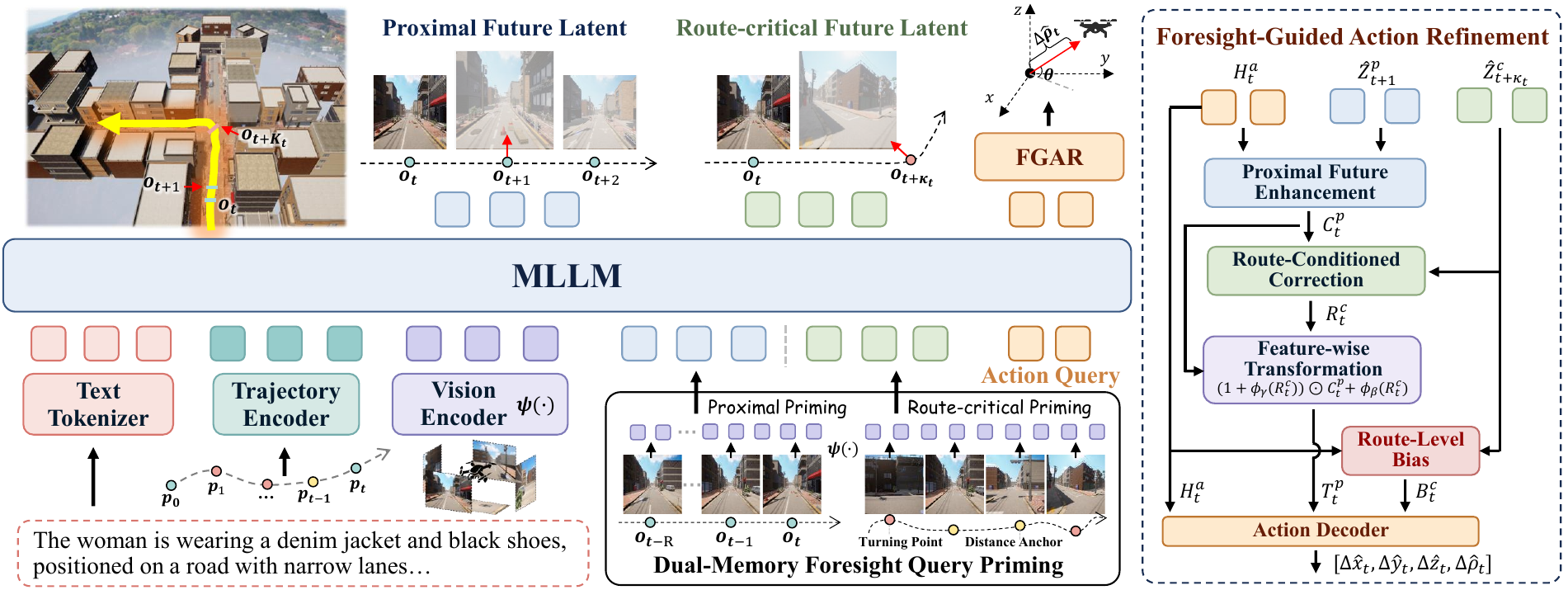}
  \caption{\textbf{Overview of ForeFly.}
    ForeFly primes dual-horizon foresight with complementary historical memories, predicts proximal and route-critical future latents, Foresight-Guided Action Refinement (FGAR) asymmetrically uses proximal foresight for local enhancement and route-critical foresight for feature-wise correction and route-level guidance.}
   \label{fig:framework}
   \vspace{-4mm}
\end{figure}

\subsection{Overview}
As illustrated in Fig.~\ref{fig:framework}, ForeFly explicitly introduces
future-world reasoning into aerial vision-language navigation instead of
directly mapping the current multimodal context to an immediate waypoint.
Given the instruction $\mathcal{I}$, trajectory history $\mathcal{P}{\leq t}$, and observation history $\mathcal{O}{\leq t}$, ForeFly leverages historical context to construct future-oriented representations and conditions waypoint prediction on the resulting dual-horizon foresight.

Specifically, ForeFly first employs \emph{Dual-Memory Foresight Query Priming} to construct recent and route-critical visual memories, providing complementary historical cues for history-aware foresight queries.
Given the primed queries, \emph{Dual-Horizon Future Latent Prediction}
jointly infers a proximal future latent and an adaptive route-critical future latent under the multimodal context.
Finally, \emph{Foresight-Guided Action Refinement} assigns distinct roles to the dual-horizon foresight, using proximal prediction for local action enhancement and route-critical prediction for route-level correction.

\subsection{Dual-Memory Foresight Query Priming}

Learnable foresight queries lack explicit route-dependent historical grounding. We therefore introduce \emph{Dual-Memory Foresight Query Priming}, which conditions horizon-specific foresight queries on complementary recent and route-critical visual memories before future latent prediction.

\noindent\textbf{Recent visual memory.}
At time step $t$, each historical observation $o_i$, $i\leq t$, is encoded as
$\mathbf{x}_i=\psi(o_i)$,
where $\psi(\cdot)$ denotes the visual encoding and projection pipeline, yielding visual embeddings $\mathbf{x}_i$.
The recent visual memory $\mathcal{M}_t^{p}$ retains features from the latest
$R$ observations, preserving short-term appearance and motion continuity around
the current state.

\noindent\textbf{Route-critical visual memory.}
While recent observations characterize local visual evolution, long-horizon
foresight benefits from sparse trajectory states associated with meaningful
route transitions.
In particular, turning events often coincide with key
navigation decision points and capture structural changes in the route
~\cite{Han_2025_CVPR, follow_2025_beaten_path}.
Let $\mathbf{p}_i\in\mathbb{R}^{3}$ denote the UAV position at time step $i$.
We define the normalized flight direction and the turning angle between
consecutive directions as:
\begin{equation}
\mathbf{v}_i =
\frac{\mathbf{p}_i-\mathbf{p}_{i-1}}
{\|\mathbf{p}_i-\mathbf{p}_{i-1}\|_2}
\in \mathbb{S}^2,
\qquad
\alpha_i =
\cos^{-1}\!\left(
\mathbf{v}_i^\top \mathbf{v}_{i+1}
\right).
\end{equation}
Frames satisfying $\alpha_i\geq\tau_p$ are selected as turning anchors,
where $\tau_p$ is the turning-angle threshold.

Turning events alone can be overly sparse along extended straight segments.
We therefore introduce distance-based coverage anchors at approximately every
$\tau_g$ meters of accumulated travel.
For two adjacent event anchors $u$ and $v$, let
$
s(u,j)=
\sum_{k=u+1}^{j}
\|\mathbf{p}_k-\mathbf{p}_{k-1}\|_2
$
denote the traveled distance from $u$ to $j$.
The $m$-th coverage anchor is selected as:
\begin{equation}
j_m^\star=
\arg\min_{j:u<j<v}
\left|
s(u,j)-m\tau_g
\right|,\
\qquad 0<m\tau_g<s(u,v).
\end{equation}
The route-critical memory $\mathcal{M}_t^{c}$ consists of visual features from the initial frame, detected turning events, and distance-based coverage anchors. This event-aware construction emphasizes route transitions while maintaining historical coverage over event-sparse trajectory segments.

\noindent\textbf{Horizon-specific foresight query priming.}
To exploit the complementary historical cues encoded in the two memories,
we introduce learnable proximal and route-critical foresight queries,
$\mathcal{Q}^{p}\in\mathbb{R}^{N_p\times d}$ and
$\mathcal{Q}^{c}\in\mathbb{R}^{N_c\times d}$, for different prediction horizons.
The proximal and route-critical queries attend to their corresponding
historical memories, $\mathcal{M}_t^{p}$ and $\mathcal{M}_t^{c}$,
yielding the history-aware queries
$\widetilde{\mathcal{Q}}_t^{p}$ and
$\widetilde{\mathcal{Q}}_t^{c}$, respectively.
The resulting history-aware queries capture recent visual continuity and route-level historical cues, respectively, and are used for dual-horizon future latent prediction.

\subsection{Dual-Horizon Future Latent Prediction}
Given the primed foresight queries, we perform dual-horizon future prediction
and action reasoning within a unified multimodal decoding process.
At time step $t$, the instruction $\mathcal{I}$, trajectory history
$\mathcal{P}_{\leq t}$, and current observation $o_t$ are encoded as
$\boldsymbol{\ell}$, $\mathbf{e}_{\leq t}^{\mathrm{traj}}$, and
$\mathbf{x}_t$, respectively.
The MLLM jointly processes these features with the primed proximal and
route-critical queries $\widetilde{\mathcal{Q}}_t^{p}$ and
$\widetilde{\mathcal{Q}}_t^{c}$, together with the learnable action queries
$\mathcal{Q}^{a}$.
Here, the primed foresight queries inject complementary historical cues from
the recent and route-critical memories, while the trajectory encoding provides
explicit spatial context.
Together, these inputs form a unified multimodal context for decoding
proximal foresight, route-critical foresight, and action representations:
\begin{equation}
\left(
\mathbf{H}_{t+1}^{p},
\mathbf{H}_{t+\kappa_t}^{c},
\mathbf{H}_{t}^{a}
\right)
=
f_{\theta}
\left(
\boldsymbol{\ell},
\mathbf{e}_{\leq t}^{\mathrm{traj}},
\mathbf{x}_t
\,;\,
\widetilde{\mathcal{Q}}_{t}^{p},
\widetilde{\mathcal{Q}}_{t}^{c},
\mathcal{Q}^{a}
\right),
\end{equation}
where $\mathbf{H}_{t+1}^{p}$, $\mathbf{H}_{t+\kappa_t}^{c}$, and
$\mathbf{H}_t^{a}$ denote the decoded representations for proximal future modeling, route-critical future modeling, and waypoint prediction, respectively. 

During this decoding process, we avoid reconstructing future RGB observations
and instead supervise the decoded foresight representations in a compact visual
embedding space. Specifically, a shared projection head $g(\cdot)$ maps the
proximal and route-critical representations to
$\hat{\mathbf{Z}}_{t+1}^{p}=g(\mathbf{H}_{t+1}^{p})$ and
$\hat{\mathbf{Z}}_{t+\kappa_t}^{c}=g(\mathbf{H}_{t+\kappa_t}^{c})$,
respectively. These auxiliary predictions are supervised by the visual
representations of the corresponding future observations, encouraging the
decoded foresight representations to capture navigation-relevant semantic
layout, scene transitions, and spatial structure without modeling unnecessary
pixel-level details. Sharing the projection head encourages the two branches
to produce future representations in a consistent latent space, while their
distinct foresight queries preserve horizon-specific temporal information.

The action-query representation $\mathbf{H}_t^{a}$ summarizes
action-oriented evidence from the current multimodal context. However, using
it directly for waypoint regression may still produce a predominantly
reactive policy.
We therefore introduce a Foresight-Guided Action Refinement (FGAR) module that explicitly refines the action representation using the proximal and route-critical foresight representations, providing complementary local and route-level guidance:
$\hat{\mathbf{a}}_t =
\Phi_{\mathrm{FGAR}}
\left(
\mathbf{H}_t^{a},
\hat{\mathbf{Z}}_{t+1}^{p},
\hat{\mathbf{Z}}_{t+\kappa_t}^{c}
\right)$,
where $\Phi_{\mathrm{FGAR}}(\cdot)$ denotes the Foresight-Guided Action
Refinement operation.

\textbf{Training Objective.}
To jointly align future-world imagination with waypoint prediction, we
supervise both the dual-horizon latent predictions and the predicted action.
The waypoint action is decomposed into a 3D direction
$\mathbf{d}_t$ and a movement magnitude $\Delta \rho_t$.
The joint training objective is:
\begin{equation}
\begin{aligned}
\mathcal{L}_{W}
&=
\mathbb{E}_{(\mathcal{C}_t,\{o_{t+\delta_h}\}_{h\in\{p,c\}})\sim\mathcal{D}}
\left[
\sum_{h\in\{p,c\}}
\frac{1}{N_h d}
\left\|
\hat{\mathbf{Z}}^{h}_{t+\delta_h}
-
\psi(o_{t+\delta_h})
\right\|_F^2
\right],
\\[1mm]
\mathcal{L}_{A}
&=
\mathbb{E}_{(\mathcal{C}_t,\mathbf{a}_t)\sim\mathcal{D}}
\left[
1-\hat{\mathbf{d}}_t^{\top}\mathbf{d}_t
+
\left|
{\Delta\hat{\rho}}_t-\Delta\rho_t
\right|
\right].
\end{aligned}
\label{eq:training_objective}
\end{equation}
where $\{\hat{\mathbf{Z}}_{t+\delta_h}^{h}\}_{h\in\{p,c\}}$ denote the predicted dual-horizon future representations at horizon $h\in\{p,c\}$, with $\delta_p=1$ and $\delta_c=\kappa_t$ denoting the proximal and
route-critical prediction horizons, respectively. 
The overall training objective is given by
$\mathcal{L}=\mathcal{L}_{W}+\mathcal{L}_{A}$.

\subsection{Foresight-Guided Action Refinement}

Within the unified multimodal decoding process, the action representation is jointly contextualized with the proximal and route-critical foresight representations.
However, this generic contextualization does not explicitly distinguish their complementary roles in action generation.
We therefore introduce a \emph{Foresight-Guided Action Refinement} (FGAR) module
that refines the action representation with dual-horizon
foresight, using proximal prediction to strengthen local action evidence and
route-critical prediction to further impose route-level guidance before
waypoint decoding.

\noindent \textbf{Proximal future enhancement.}
The proximal future latent describes the immediate visual evolution from the
current observation. We first use the action representation as the query and
the predicted proximal latent as the key and value
$
\mathbf{C}_t^{p}
=
\operatorname{Attn}_{p}
(
\mathbf{H}_t^{a},
\hat{\mathbf{Z}}_{t+1}^{p}
),
$
where $\mathbf{C}_t^{p}$ denotes the proximal-enhanced action representation.
This interaction grounds the waypoint decision in anticipated short-term
changes in viewpoint, local geometry, and scene layout.

\noindent \textbf{Asymmetric route-critical conditioning.}
A straightforward fusion of the proximal and route-critical future latents
would treat the two horizons as equally relevant to action prediction.
However, they play fundamentally different roles in aerial navigation.
The proximal future describes locally executable scene evolution, whereas the
route-critical future provides sparse but strategically important information
about upcoming turns, landmark transitions, and route progress. We therefore
adopt an asymmetric conditioning strategy: the proximal-enhanced action
representation serves as the execution carrier, while the route-critical
future acts as a high-level controller that selectively reshapes this local
action evidence.

Specifically, we first derive a route-conditioned correction context by
allowing the proximal-enhanced action representation to attend to the
route-critical future latent
$
\mathbf{R}_t^{c}
=
\operatorname{Attn}_{c}
(
\mathbf{C}_t^{p},
\hat{\mathbf{Z}}_{t+\kappa_t}^{c}
).
$
Rather than directly decoding $\mathbf{R}_t^{c}$ into an action or replacing
the locally grounded representation $\mathbf{C}_t^{p}$, we use it to generate
a feature-wise transformation:
\begin{equation}
\left[
\boldsymbol{\gamma}_t,
\boldsymbol{\beta}_t
\right]
=
\phi_{\mathrm{film}}
\left(
\mathbf{R}_t^{c}
\right),
\qquad
\mathbf{T}_t^{p}
=
\left(
1+\boldsymbol{\gamma}_t
\right)
\odot
\mathbf{C}_t^{p}
+
\boldsymbol{\beta}_t.
\end{equation}
Here, $\mathbf{T}_t^{p}$ denotes the route-corrected proximal action
representation. The scaling factor $\boldsymbol{\gamma}_t$ controls which
dimensions of the local action evidence should be amplified or suppressed,
while $\boldsymbol{\beta}_t$ introduces event-dependent feature shifts.
Consequently, the route-critical future modifies the interpretation of the
proximal future without discarding its locally executable geometric
information.

Since feature-wise modulation may overlook route-level cues,
we introduce a parallel
route-level guidance path. The original action representation directly queries the
route-critical future latent:
\begin{equation}
\mathbf{B}_t^{c}
=
\phi_{\mathrm{bias}}
\left(
\operatorname{Attn}_{b}
\left(
\operatorname{sg}\!\left(\mathbf{H}_t^{a}\right),
\operatorname{sg}\!\left(
\hat{\mathbf{Z}}_{t+\kappa_t}^{c}
\right)
\right)
\right),
\end{equation}
where $\operatorname{sg}(\cdot)$ denotes the stop-gradient operation and $\phi_{\mathrm{bias}}(\cdot)$ denotes the learnable projection. Unlike the modulation path, this branch produces an additive route-level decision $\mathbf{B}_t^{c}$. Detaching its inputs makes the branch a read-only consumer, preventing the additive path from becoming an optimization shortcut. Meanwhile, the main modulation path remains differentiable, allowing the route-critical representation to be jointly optimized for future prediction and downstream action utility.

The final action integrates the original action evidence, route-corrected proximal features, and the detached route-level guidance residual:
\begin{equation}
\hat{\mathbf{a}}_t
=
\phi_{\mathrm{act}}
\left(
\operatorname{LN}
\left(
\mathbf{H}_t^{a}
+
\mathbf{T}_t^{p}
+
\mathbf{B}_t^{c}
\right)
\right).
\end{equation}
The original representation $\mathbf{H}_t^{a}$ acts as a residual basis,
$\mathbf{T}_t^{p}$ preserves local executability while incorporating
event-conditioned feature correction, and $\mathbf{B}_t^{c}$ provides
additive route-level guidance. The refined representation is then
decoded into the waypoint action through the MLP projection $\phi_{\mathrm{act}}(\cdot)$.

\section{Experiment}

\subsection{Experimental Setup}
\textbf{Datasets.}
We conduct experiments on two widely used AVLN benchmarks, TravelUAV~\cite{wang2024openuav} and UAV-ON~\cite{uav-on}. 
TravelUAV contains 12,149 trajectories across 22 diverse scenarios, with target distances ranging from 50\,m to 400\,m.
UAV-ON further provides 14 high-fidelity outdoor environments with diverse geographic structures, ranging from compact urban parks to expansive mountainous terrains. 
Together, these two benchmarks enable a comprehensive evaluation of navigation robustness and generalization across diverse aerial environments.

\noindent \textbf{Evaluation Metrics.} 
Navigation performance is evaluated using four standard VLN metrics: Navigation Error (NE), Success Rate (SR), Oracle Success Rate (OSR), and Success weighted by Path Length (SPL). NE measures the final distance to the goal, SR reports the success rate of reaching the goal, and OSR checks whether the goal region is reached at any point along the trajectory. SPL additionally evaluates path efficiency by considering the ratio between the shortest and executed path lengths, providing a balanced assessment of navigation success and efficiency.

\noindent\textbf{Implementation Details.}
All experiments are conducted using PyTorch on four NVIDIA RTX 5090 GPUs. We adopt LLaMA 2-7B as the language backbone and EVA-ViT-G as the visual encoder. The numbers of proximal and route-critical foresight slots are both set to $N_p=N_c=17$. The turning-angle threshold $\tau_p$ and distance interval $\tau_g$ are set to $20^\circ$ and $30$~m, respectively. We fine-tune the model with LoRA using a learning rate of $5\times10^{-5}$. 
Training proceeds in two stages. In the first stage, we replace FGAR with a simple MLP head to stabilize future representation learning and waypoint prediction. In the second stage, FGAR is restored and all trainable components are jointly optimized, enabling future representations to effectively guide navigation decisions.

\subsection{Quantitative Results}
\begin{table}[t]
\caption{Results on the TravelUAV Unseen dataset. NE is reported in meters (m), while SR, OSR, and SPL are reported as percentages (\%). For task difficulty, Easy denotes trajectories shorter than 250 meters, whereas Hard denotes those longer than 250 meters.
The best results are bold.}
\label{tab:openuav_unseen_results}
\centering
\small
\setlength{\tabcolsep}{7pt}
\renewcommand{\topfraction}{0.9}
\renewcommand{\textfraction}{0.05}
\renewcommand{\floatpagefraction}{0.8}
\begin{adjustbox}{max width=\linewidth}
\begin{tabular}{lcccccccccccc}
\toprule
\multirow{2}{*}{Method} &
\multicolumn{4}{c}{\textbf{Full}} &
\multicolumn{4}{c}{\textbf{Easy}} &
\multicolumn{4}{c}{\textbf{Hard}} \\
\cmidrule(lr){2-5} \cmidrule(lr){6-9} \cmidrule(lr){10-13}
& NE $\!\downarrow$ & SR $\!\uparrow$ & OSR $\!\uparrow$ & SPL $\!\uparrow$
& NE $\!\downarrow$ & SR $\!\uparrow$ & OSR $\!\uparrow$ & SPL $\!\uparrow$
& NE $\!\downarrow$ & SR $\!\uparrow$ & OSR $\!\uparrow$ & SPL $\!\uparrow$ \\
\midrule
Random Action               & 225.64 & 0.06  & 0.06  & 0.06  & 164.66 & 0.19  & 0.19  & 0.19  & 280.58 & 0.00  & 0.00  & 0.00 \\
Fixed Action                & 193.30 & 1.76  & 5.36  & 1.09  & 140.33 & 3.19  & 8.08  & 1.88  & 245.96 & 0.85  & 3.08  & 0.55 \\
CMA~\cite{anderson2017r2r}                  & 147.27 & 4.98  & 12.41 & 4.74  & 102.54 & 8.03  & 17.52 & 7.52  & 191.30 & 2.76  & 7.53  & 2.71 \\
TravelUAV~\cite{wang2024openuav}              & 130.60 & 11.41 & 31.13 & 10.45 & 96.27  & 12.47 & 33.31 & 11.29 & 167.49 & 10.62 & 28.91 & 9.80 \\
NavFoM~\cite{zhang2025embodied}	& 118.34	& 15.63	& 30.46	& 14.21	& 89.77	& 16.98	& 32.22	& 15.35	& 155.69	& 14.35	& 27.79	& 13.16 \\

LongFly~\cite{jiang2025longfly} &91.84 &24.19 &43.86 &20.84 &69.16 &22.89 &43.24 &18.66 &112.02 &25.36 &44.41 &22.76\\

SpatialFly~\cite{jiang2026spatialfly}      
                     & 87.82  & 25.46 & 44.41 & 21.76
                     & 64.49  & 25.17 & 44.18 & 20.36
                     & 108.56 & 25.71 & 44.63 & 23.01 \\

\rowcolor{blue!15}
\textbf{ForeFly (Ours)} 
&\textbf{79.80}	&\textbf{28.86}	&\textbf{52.87}	&\textbf{25.19}
&\textbf{62.07}	&\textbf{26.49}	&\textbf{53.25}	&\textbf{22.70}
&\textbf{96.51}	&\textbf{31.09}	&\textbf{52.51}	&\textbf{27.54} \\

\bottomrule
\end{tabular}
\end{adjustbox}
\end{table}

\begin{table}[t]
\caption{Results on the TravelUAV Seen dataset. \emph{Human} represents the performance of human operators under manual UAV control.
The best results are bold.}
\label{tab:openuav_seen_results}
\centering
\small
\setlength{\tabcolsep}{6.5pt}
\renewcommand{\arraystretch}{1.12}

\begin{adjustbox}{max width=\linewidth}
\begin{tabular}{lcccccccccccc}
\toprule
\multirow{2}{*}{\textbf{Method}} &
\multicolumn{4}{c}{\textbf{Full}} &
\multicolumn{4}{c}{\textbf{Easy}} &
\multicolumn{4}{c}{\textbf{Hard}} \\
\cmidrule(lr){2-5} \cmidrule(lr){6-9} \cmidrule(lr){10-13}
& NE $\!\downarrow$ & SR $\!\uparrow$ & OSR $\!\uparrow$ & SPL $\!\uparrow$
& NE $\!\downarrow$ & SR $\!\uparrow$ & OSR $\!\uparrow$ & SPL $\!\uparrow$
& NE $\!\downarrow$ & SR $\!\uparrow$ & OSR $\!\uparrow$ & SPL $\!\uparrow$ \\
\midrule

\rowcolor{gray!20}
Human & \textbf{14.15} & \textbf{94.51} & \textbf{94.51} & \textbf{77.84}
      & \textbf{11.68} & \textbf{95.44} & \textbf{95.44} & \textbf{76.19}
      & \textbf{17.16} & \textbf{93.37} & \textbf{93.37} & \textbf{79.85} \\
\midrule

Random Action   &222.20 &0.14 &0.21 &0.07 
                &142.07 &0.26 &0.39 &0.13 
                &320.12 &0.00 &0.00 &0.00 \\
Fixed Action    &188.61 &2.27 &8.16 &1.40 
                &121.36 &3.48 &11.48 &2.14 
                &270.69 &0.79 &4.09 &0.49 \\
CMA~\cite{anderson2017r2r} 
                &135.73 &8.37 &18.72 &7.90 
                &84.89 &11.48 &24.52 &10.68 
                &197.77 &4.57 &11.65 &4.51 \\

TravelUAV~\cite{wang2024openuav}
               & 106.28 & 16.10 & 44.26 & 14.30
               & 68.78  & 18.84 & 47.61 & 16.39
               & 152.04 & 12.76 & 40.16 & 11.76 \\

TravelUAV-DA~\cite{wang2024openuav}
               & 98.66  & 17.45 & 48.87 & 15.76
               & 66.40  & 20.26 & 51.23 & 18.10
               & 138.04 & 14.02 & 45.98 & 12.90 \\

SimpleNav~\cite{simplenav}
                & 85.60 & 22.40 & 55.10 & 20.50
                & 60.00 & 22.80 & 56.90 & 21.00
                & 118.30 & 22.00 & 52.80 & 19.90 \\
NavFoM~\cite{zhang2025embodied}
               & 93.05 & 29.17 & 49.24 & 25.03
               & 58.98 & 32.91 & 53.16 & 27.87
               & 143.83 & 23.58 & 43.40 & 20.80 \\
AirForsight~\cite{liu2026airforesight}
                & 56.99 & 35.83 & 69.25 & 30.22
                & 38.27 & 38.83 & 71.26 & 31.65
                & 80.08 & 32.13 & 66.77 & 28.45 \\
LongFly~\cite{jiang2025longfly} 
        & 60.02 & 36.39 & 65.87 & 31.07 
        & 38.10 & 38.52 & 71.90 & 31.24 
        & 85.20 & 33.94 & 58.94 & 30.88 \\
SpatialFly~\cite{jiang2026spatialfly}
&58.05	&38.54	&68.62	&33.43
&\textbf{34.41}	&39.59	&\textbf{74.14}	&33.61
&84.76	&37.33	&62.27	&33.22 \\

\rowcolor{red!15}
\textbf{ForeFly (Ours)}
&\textbf{48.74}	&\textbf{41.75}	&\textbf{70.25}	&\textbf{34.75}
&34.49	&\textbf{44.55}	&71.36	&\textbf{35.72}
&\textbf{66.16}	&\textbf{38.33}	&\textbf{68.89}	&\textbf{33.55} \\

\bottomrule

\end{tabular}
\end{adjustbox}
\end{table}

\textbf{Results on TravelUAV Dataset.}
As shown in Tables~\ref{tab:openuav_unseen_results} and~\ref{tab:openuav_seen_results}, ForeFly consistently achieves strong performance on both the Unseen and Seen splits.
On the Unseen split, ForeFly reduces NE by 8.02~m and improves SR, OSR, and SPL by 3.40, 8.46, and 3.43 percentage points over SpatialFly, respectively, indicating improved generalization to unseen environments.
The advantage is particularly pronounced on the Hard subset, where SR increases from 25.71\% to 31.09\%, highlighting its effectiveness on long-range navigation tasks.
On the Seen split, ForeFly achieves the best overall performance among autonomous methods. Compared with SpatialFly, ForeFly improves SR by 3.21 and 4.96 percentage points on the Full and Easy subsets, respectively.
Although its OSR on the Easy subset is slightly lower than that of SpatialFly, ForeFly improves OSR by 6.62 percentage points on the Hard subset, indicating a clearer advantage on more challenging long-range trajectories.

\textbf{Results on UAV-ON Dataset.}
As shown in Table~\ref{tab:uavon_results}, ForeFly achieves the best performance across all metrics on both the Seen and Unseen splits. On the Seen split, ForeFly reaches 25.13\% SR, 36.30\% OSR, and 16.31\% SPL, outperforming the strongest competing results by 2.41, 5.82, and 7.18 percentage points, respectively. On the Unseen split, ForeFly improves SR from 17.57\% to 22.68\%, yielding a 5.11 percentage-point gain, while also achieving the highest OSR of 31.44\% and SPL of 14.59\%. These consistent improvements across both splits further demonstrate the robustness and generalization ability of ForeFly beyond the TravelUAV benchmark.

\begin{table}[t]
\caption{Ablation study of the main components in ForeFly. Seen and Unseen represent the \textit{Test-seen} and \textit{Test-unseen sets}, respectively.}
\label{tab:component_ablation}
\centering
\small
\setlength{\tabcolsep}{4pt}
\renewcommand{\arraystretch}{1.15}
\begin{adjustbox}{max width=\linewidth}
\begin{tabular}{lcccccccccccccccc}
\toprule
\multirow{2}{*}{Variant}
& \multirow{2}{*}{\makecell[c]{Proximal\\Foresight}}
& \multirow{2}{*}{\makecell[c]{Route-Critical\\Foresight}}
& \multirow{2}{*}{\makecell[c]{Foresight Query\\Priming}}
& \multirow{2}{*}{\makecell[c]{Foresight-Guided\\Action Refinement}}
& \multicolumn{4}{c}{\textbf{Seen (Full)}}
& \multicolumn{4}{c}{\textbf{Unseen (Full)}} \\
\cmidrule(lr){6-9} \cmidrule(lr){10-13} \cmidrule(lr){14-17}
& & & & 
& NE $\downarrow$ & SR $\uparrow$ & OSR $\uparrow$ & SPL $\uparrow$
& NE $\downarrow$ & SR $\uparrow$ & OSR $\uparrow$ & SPL $\uparrow$ \\
\midrule

Baseline
& $\times$ & $\times$ & $\times$ & $\times$
& 103.75 & 21.73 & 42.41 & 18.04
& 130.60 & 11.41 & 31.13 & 10.45 \\

Proximal Only
& $\checkmark$ & $\times$ & $\times$ & $\times$
& 72.16 & 23.00 & 54.50 & 19.46
& 84.50 & 19.25 & 44.25 & 16.91 \\

Route-Critical Only
& $\times$ & $\checkmark$ & $\times$ & $\times$
& 69.99 & 28.25 & 57.25 & 24.19
& 95.00 & 17.50 & 42.00 & 15.37 \\

Dual-Horizon 
& $\checkmark$ & $\checkmark$ & $\times$ & $\times$
& 70.06  & 32.75 & 62.00 & 27.54
& 84.14  & 21.75 & 44.53 & 16.29 \\

+ Foresight Query Priming
& $\checkmark$ & $\checkmark$ & $\checkmark$ & $\times$
& 64.41  & 33.08 & 62.91 & 27.99
& 81.82  & 23.61 & 45.36 & 19.73 \\

\rowcolor{blue!10}
\textbf{ForeFly (Full)}
& $\checkmark$ & $\checkmark$ & $\checkmark$ & $\checkmark$
&\textbf{48.74}	&\textbf{41.75}	&\textbf{70.25}	&\textbf{34.75}
&\textbf{79.80}	&\textbf{28.86}	&\textbf{52.87}	&\textbf{25.19} \\
\bottomrule
\end{tabular}
\end{adjustbox}
\vspace{-2mm}
\end{table}

\begin{table}[htbp]
\vspace{-2mm}
\centering
\begin{minipage}[c]{0.55\textwidth}
\centering
\caption{Results on the UAV-ON dataset.}
\label{tab:uavon_results}
\vspace{2pt} %
\resizebox{\linewidth}{!}{%
\setlength{\tabcolsep}{4pt}
\renewcommand{\arraystretch}{1.15}
\begin{tabular}{lcccccc}
\toprule
\multirow{2}{*}{\textbf{Method}} & \multicolumn{3}{c}{\textbf{Seen}} & \multicolumn{3}{c}{\textbf{Unseen}} \\
\cmidrule(lr){2-4} \cmidrule(lr){5-7}
 & \textbf{SR$\uparrow$} & \textbf{OSR$\uparrow$} & \textbf{SPL$\uparrow$} & \textbf{SR$\uparrow$} & \textbf{OSR$\uparrow$} & \textbf{SPL$\uparrow$} \\
\midrule
AerialVLN~\cite{aerialvln} & 4.55 & 16.84 & 4.24 & 3.51 & 20.77 & 3.41 \\
AOA-F~\cite{uav-on} & 7.87 & 17.98 & 4.21 & 6.45 & 16.09 & 3.98 \\
OpenFly~\cite{gao2026openfly} & 12.29 & 26.20 & 6.83 & 11.82 & 25.71 & 5.64 \\
APEX~\cite{apex} & 12.36 & 19.10 & 9.13 & 16.13 & 22.58 & 13.03 \\
Navid~\cite{navid} & 14.17 & 30.08 & 7.19 & 10.70 & 28.51 & 5.99 \\
OctMem-Agent~\cite{octmem} & 22.72 & 30.48 & 8.40 & 17.57 & 28.59 & 5.15 \\
\rowcolor{blue!15}
\textbf{ForeFly (Ours)} & \textbf{25.13} & \textbf{36.30} & \textbf{16.31} & \textbf{22.68} & \textbf{31.44} & \textbf{14.59} \\
\bottomrule
\end{tabular}%
}
\end{minipage}
\hfill %
\begin{minipage}[c]{0.42\textwidth}
\centering
\caption{Ablation on Foresight-Guided Action Refinement (FGAR) Design. }
\label{tab:FGAR_ablation}
\vspace{2pt}
\resizebox{\linewidth}{!}{%
\setlength{\tabcolsep}{4pt}
\renewcommand{\arraystretch}{1.15}
\begin{tabular}{lcccc}
\toprule
\textbf{Method} & \textbf{NE$\downarrow$} & \textbf{SR$\uparrow$} & \textbf{OSR$\uparrow$} & \textbf{SPL$\uparrow$} \\
\midrule
MLP Head & 64.41 & 33.08 & 62.91 & 27.99 \\
Direct Dual-Future* & 62.03 & 34.00 & 62.50 & 28.73 \\
FGAR \textit{w/o} $\mathbf{B}_t^{c}$ & 54.61 & 38.35 & 68.73 & 32.55 \\
\rowcolor{blue!10}
\textbf{FGAR} & \textbf{48.74} & \textbf{41.75} & \textbf{70.25} & \textbf{34.75} \\
\bottomrule
\end{tabular}%
}
\par\vspace{3pt}
\parbox{\linewidth}{%
\scriptsize
\textit{Note:} \emph{Direct Dual-Future Fusion} refers to concatenating
the proximal and route-critical future latents and using them to
condition the action representation via \emph{cross-attention}.
}
\end{minipage}

\end{table}

Overall, the consistent gains across both TravelUAV and UAV-ON, particularly on unseen environments and challenging long-range trajectories, demonstrate the robustness and generalization of ForeFly for long-horizon AVLN. These improvements are consistent with the complementary dual-horizon foresight design, where proximal prediction preserves local motion continuity, while route-critical foresight provides strategic route-level anticipation. By integrating these future cues into waypoint generation through FGAR, ForeFly can reduce accumulated navigation errors and maintain better route consistency over extended trajectories.

\subsection{Ablation study}
\label{sec:ablation}
As shown in Table~\ref{tab:component_ablation}, we conduct a systematic ablation study to evaluate the contribution of each key component in ForeFly. 
We progressively examine the effects of proximal foresight, route-critical foresight, Dual-Memory Foresight Query Priming, and the Foresight-Guided Action Refinement on TravelUAV dataset, isolating their individual and complementary contributions.

\textbf{Effect of Dual-Horizon Foresight.}
We first examine the contribution of the two foresight horizons. Both Proximal Only and Route-Critical Only consistently outperform the baseline, confirming the benefit of explicitly modeling future states. Interestingly, Route-Critical Only yields larger gains on the Seen split, whereas Proximal Only performs better on the Unseen split. Combining the two horizons further improves SR to 32.75\% and 21.75\% on the Seen and Unseen splits, respectively. These results suggest that proximal and route-critical foresight provide complementary information at different navigation horizons rather than redundant predictions.

\textbf{Effect of Foresight Query Priming.}
We next examine whether grounding foresight queries in historical observations improves future-state modeling. Introducing retrospective priming consistently benefits both splits, with larger gains on the Unseen split, where SR increases from 21.75\% to 23.61\% and SPL from 16.29\% to 19.73\%. These results suggest that route-dependent historical cues enrich future-state prediction and improve foresight reliability in unseen environments.

\textbf{Effect of Foresight-Guided Action Refinement.}
Finally, we evaluate the effectiveness of incorporating predicted future representations into navigation decisions. Introducing FGAR substantially improves the retrospectively primed dual-horizon model, reducing NE from 64.41~m to 48.74~m and increasing SR from 33.08\% to 41.75\% on the Seen split. A similar improvement is observed on the Unseen split, where SR increases from 23.61\% to 28.86\%. 
The consistent gains across both splits confirm the effectiveness of future representations for action refinement.
These results show that future-state prediction alone is insufficient, effectively conditioning action generation on the predicted futures is crucial for translating foresight into reliable navigation decisions.

\textbf{Ablation on Foresight-Guided Action Refinement Design.}
We further compare different strategies for incorporating dual-horizon foresight into action refinement. 
As shown in Table~\ref{tab:FGAR_ablation}, directly fusing the two future latents yields only modest improvements over the MLP head, suggesting that simply conditioning the action representation on future information is insufficient.
In contrast, the asymmetric conditioning mechanism substantially improves performance, and the additional route-level guidance branch further strengthens all metrics.
This validates the design choice of assigning different roles to proximal and route-critical foresight during action refinement.

\subsection{Qualitative Analysis}

\begin{figure}[!t]
    \centering
    \includegraphics[width=0.98\linewidth]{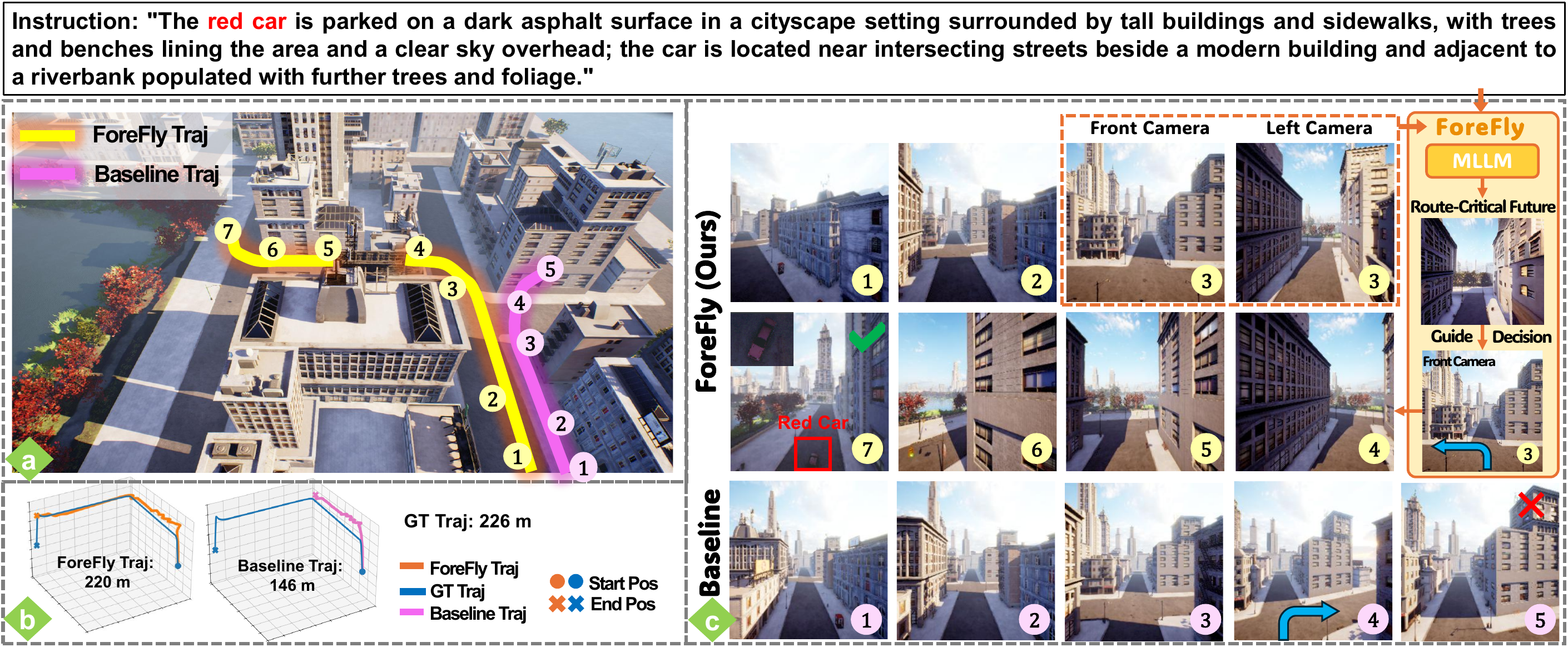}
    \vspace{-3mm}
    \caption{\textbf{Qualitative Results.}
    (a) Bird’s-eye view of predicted trajectories. 
    (b) 3D trajectory comparison showing improved route consistency of ForeFly. 
    (c) Step-by-step navigation, where route-critical foresight helps ForeFly anticipate the upcoming turn, while the baseline deviates.
    The corresponding future visualization procedure is described in Sec.~\ref{sec:qual_future}.
    \vspace{-4mm}
    }
    \label{fig:qualitative_analysis}
\end{figure}

Figure~\ref{fig:qualitative_analysis} presents a representative example comparing ForeFly with the baseline.
As illustrated in Fig.~\ref{fig:qualitative_analysis}(a), the UAV is instructed to locate a red car near a riverbank after traversing the urban blocks.
As shown in Fig.~\ref{fig:qualitative_analysis}(c), the baseline makes an incorrect decision at the critical turning point and consequently deviates from the intended route.
In contrast, ForeFly internally anticipates a route-critical future latent associated with the upcoming route transition, providing strategic guidance for the waypoint decision and enabling the UAV to make the correct turn and ultimately reach the target.

The trajectory comparisons in Fig.~\ref{fig:qualitative_analysis}(a,b) further illustrate this advantage.
ForeFly generates a 220~m trajectory that closely follows the 226~m ground-truth route, while the baseline exhibits an early route deviation and fails to complete the navigation.
These results highlight the complementary roles of the two foresight horizons: proximal foresight supports locally consistent waypoint generation, whereas route-critical foresight provides strategic guidance around important landmarks and turning regions. Together, they enable ForeFly to maintain both local motion continuity and long-horizon route consistency in complex aerial environments. 

\section{Conclusion}

In this work, we introduced ForeFly, a dual-horizon latent world action model for aerial vision-language navigation. 
Rather than relying solely on reactive action prediction or single-horizon foresight, ForeFly explicitly anticipates both proximal future states for local motion continuity and route-critical future states for long-range guidance. 
A foresight query priming mechanism grounds these predictions in complementary historical cues, while the proposed Foresight-Guided Action Refinement (FGAR) asymmetrically integrates the two foresight horizons to preserve locally executable decisions while enabling strategic route-level correction. 
Experiments on the TravelUAV and UAV-ON benchmarks demonstrate consistent improvements over strong baselines on both seen and unseen environments, with particularly clear gains on challenging long-horizon trajectories. Ablation and qualitative analyses further validate the complementary roles of dual-horizon foresight, foresight query priming, and foresight-guided action refinement.

\bibliography{main}
\bibliographystyle{conference}

\appendix
\clearpage
\section{Implementation Details}
\label{sec:implementation}

\subsection{Route-Critical Anchors and Future Targets}
\label{sec:anchor_target}

\begin{figure}[t]
  \centering
  \includegraphics[width=\textwidth]{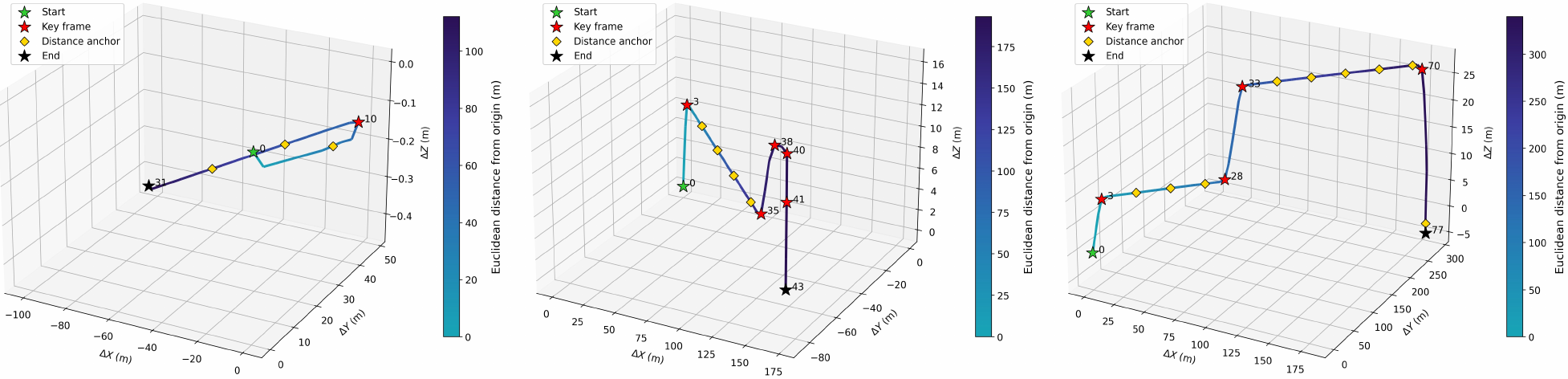}
  \caption{\textbf{Visualization of route-critical anchor construction.}
        Representative UAV trajectories of different lengths are shown. Green and black stars mark the start and end states, red stars indicate key frames at turning points, and yellow diamonds denote distance anchors. The color scale represents the Euclidean distance from the starting position.}
   \label{fig:key_frame}
\end{figure}

\paragraph{Anchor construction.}
Let a demonstration contain $T$ states and let
$\mathbf{p}_i\in\mathbb{R}^{3}$ be the UAV position at state $i$.  We first
form an event-anchor set from the initial state, the terminal state, and every
turning point whose angular change satisfies $\alpha_i\geq\tau_p$:
\begin{equation}
    \mathcal{E}
    =
    \{1,T\}
    \cup
    \left\{i\mid \alpha_i\geq\tau_p\right\}.
\end{equation}
The initial state makes the construction well defined from the beginning of a
trajectory, while the terminal state guarantees a future target for every
non-terminal training state.  Turning points alone can be sparse on a long
straight segment.  Therefore, for each pair of consecutive event anchors
$u<v$, we add the state whose accumulated travel distance from $u$ is closest
to each multiple of $\tau_g$ that lies strictly inside the interval:
\begin{equation}
\begin{aligned}
    s(u,j)
    &=
    \sum_{k=u+1}^{j}
    \left\|\mathbf{p}_{k}-\mathbf{p}_{k-1}\right\|_2,\\
    j_m^{\star}
    &=
    \operatorname*{arg\,min}_{u<j<v}
    \left|s(u,j)-m\tau_g\right|,
    \qquad 0<m\tau_g<s(u,v).
\end{aligned}
\end{equation}
Ties are resolved by choosing the earlier state, and duplicate indices caused
by discrete sampling are removed.  Sorting the union of event and coverage
anchors gives the ordered set
\begin{equation}
    \mathcal{A}
    =
    \operatorname{sort}
    \left(
        \mathcal{E}
        \cup
        \{j_m^{\star}\}_{u,v,m}
    \right).
\end{equation}
This construction preserves turning events while preventing the target
horizon from becoming excessively long in event-sparse portions of a route.
As illustrated in Fig.~\ref{fig:key_frame}, the resulting anchors capture
turning events while maintaining coverage along trajectories of different
lengths.
In all experiments, $\tau_p=20^{\circ}$ and $\tau_g=30$~m.

\paragraph{Adaptive future target.}
For every non-terminal training state $t<T$, the next route-critical anchor is:
\begin{equation}
    a_t^{+}
    =
    \min\left\{a\in\mathcal{A}\mid a>t\right\},
    \qquad
    \kappa_t=a_t^{+}-t.
    \label{eq:supp_next_anchor}
\end{equation}
The two visual targets are consequently:
\begin{equation}
    \mathbf{Z}_{t+1}^{p,*}=\psi(o_{t+1}),
    \qquad
    \mathbf{Z}_{t+\kappa_t}^{c,*}
    =\psi(o_{a_t^{+}}).
    \label{eq:supp_dual_targets}
\end{equation}
The proximal branch is always trained against the immediately following
observation.  In contrast, the route-critical branch is trained against the
next turning, coverage, or terminal anchor, so its temporal offset adapts to
the geometry of the demonstrated route.  Importantly, $\kappa_t$ is a label
construction variable rather than an input to the model.  The notation
$t+\kappa_t$ describes which future state provides supervision; ForeFly is not
given this offset at test time.

\subsection{Multi-View Perception and Historical Memory}
\label{sec:supp_multiview_memory}

At each navigation step, the UAV receives five synchronized egocentric views,
including front, rear, left, right, and down:
\begin{equation}
    o_t =
    \left\{
    o_t^{\mathrm{front}},
    o_t^{\mathrm{rear}},
    o_t^{\mathrm{left}},
    o_t^{\mathrm{right}},
    o_t^{\mathrm{down}}
    \right\}.
\end{equation}
All five views are used for current perception, while historical memory stores
only the front-view features. For each historical frame, the visual encoder
produces one global token and 16 local tokens,
$\mathbf{x}_i^{\mathrm{front}}
=[\mathbf{g}_i;\mathbf{l}_{i,1},\ldots,\mathbf{l}_{i,16}]$.
The recent memory $\mathcal{M}_t^{p}$ retains all 17 tokens from the latest
$R=5$ front-view observations. Keeping both global and local tokens preserves
the fine-grained appearance and spatial details needed to model short-term
visual continuity. In contrast, the route-critical memory
$\mathcal{M}_t^{c}$ stores only the global token $\mathbf{g}_i$ from each
selected historical anchor, since its purpose is to summarize sparse
route-level events rather than local visual details.

Both memories are constructed from the executed trajectory prefix. 
In particular, a turning-point anchor becomes available only after the
subsequent state required to determine its turning angle has been observed,
and coverage anchors are likewise identified only from past executed states.
Future anchors may therefore be used to define training targets, but never
enter the model input before they become observable.

At inference time, ForeFly is conditioned only on:
\begin{equation}
    \mathcal{C}_t^{\mathrm{in}}
    =
    \left(
    \mathcal{I},
    \mathcal{P}_{\leq t},
    o_t,
    \mathcal{M}_t^{p},
    \mathcal{M}_t^{c}
    \right),
\end{equation}
and does not access future observations, future anchor indices, or the adaptive
offset $\kappa_t$, thereby preventing target leakage.

\subsection{Online Memory Update}
\label{sec:online_memory_update}
Algorithm~\ref{alg:online_memory_update} describes the online memory update procedure.
After observing state $t$, only turns at indices $i\leq t-1$ can be
confirmed. The current index $t$ serves as a temporary coverage boundary,
not as a known terminal anchor. Thus, distance anchors can be selected
before the next turn is observed. Let $D_i$ denote cumulative travel
distance from the initial state, so that $s(u,j)=D_j-D_u$.

\begin{algorithm}[H]
\caption{Online update of recent and route-critical memories}
\label{alg:online_memory_update}
\small
\begin{algorithmic}[1]
\Require Observed position $\mathbf{p}_t$ and front-view tokens
  $\mathbf{x}_t^{\mathrm{front}}=[\mathbf{g}_t;\mathbf{l}_{t,1},\ldots,\mathbf{l}_{t,16}]$
\Require Recent window $R=5$, turning threshold $\tau_p=20^\circ$,
  coverage interval $\tau_g=30$\,m
\State Maintain confirmed event indices $\mathcal{E}$ and cached past global tokens
\State Cache $\mathbf{g}_t$; append $\mathbf{x}_t^{\mathrm{front}}$ to the recent FIFO queue
\State Retain the latest $\min(R,t)$ entries as $\mathcal{M}_t^p$
\If{$t=1$}
  \State $D_1\gets 0$; $\mathcal{E}\gets\{1\}$
\Else
  \State $D_t\gets D_{t-1}+\|\mathbf{p}_t-\mathbf{p}_{t-1}\|_2$
\EndIf
\If{$t\geq 3$ and both adjacent displacements are nonzero}
  \State $\mathbf{v}_{t-1}\gets(\mathbf{p}_{t-1}-\mathbf{p}_{t-2})/\|\mathbf{p}_{t-1}-\mathbf{p}_{t-2}\|_2$
  \State $\mathbf{v}_{t}\gets(\mathbf{p}_{t}-\mathbf{p}_{t-1})/\|\mathbf{p}_{t}-\mathbf{p}_{t-1}\|_2$
  \State $\alpha_{t-1}\gets\arccos(\operatorname{clip}(\mathbf{v}_{t-1}^{\top}\mathbf{v}_t,-1,1))$
  \If{$\alpha_{t-1}\geq\tau_p$}
    \State $\mathcal{E}\gets\mathcal{E}\cup\{t-1\}$ \Comment{Confirm the previous state as a turn}
  \EndIf
\EndIf
\State $\mathcal{B}_t\gets\operatorname{sort}(\mathcal{E}\cup\{t\})$;
  $\mathcal{A}_t^{\mathrm{mem}}\gets\mathcal{E}$
\For{each consecutive pair $(u,v)$ in $\mathcal{B}_t$ with $v>u+1$}
  \For{each positive integer $m$ satisfying $m\tau_g<D_v-D_u$}
    \State $j^\star\gets\operatorname*{arg\,min}_{u<j<v}|D_j-D_u-m\tau_g|$
      \Comment{Break ties toward earlier $j$}
    \State $\mathcal{A}_t^{\mathrm{mem}}\gets\mathcal{A}_t^{\mathrm{mem}}\cup\{j^\star\}$
  \EndFor
\EndFor
\State $\mathcal{M}_t^c\gets[\mathbf{g}_i]_{i\in\operatorname{sort}(\mathcal{A}_t^{\mathrm{mem}})}$
\State \Return $\mathcal{M}_t^p,\mathcal{M}_t^c$ for foresight query priming at step $t$
\end{algorithmic}
\end{algorithm}

\section{Additional Experiment Results}
\subsection{Quantitative Results on Unseen Map and Unseen Object Subsets}
\label{sec:unseen_map_object}

Following the original TravelUAV evaluation protocol, the Unseen split consists of two subsets: Unseen Map and Unseen Object. The Unseen Map subset contains trajectories in previously unseen environments, while the Unseen Object subset contains trajectories with unseen goal objects.

\begin{figure}[t]
  \centering
  \includegraphics[width=\textwidth]{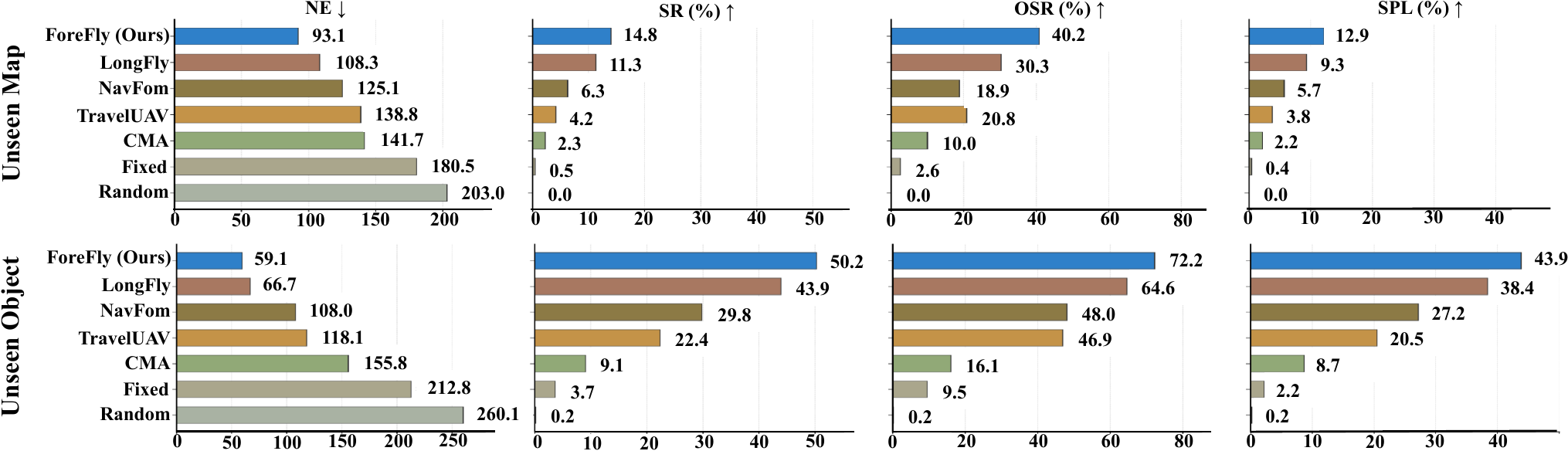}
  \caption{\textbf{Results on TravelUAV Unseen Map and Unseen Object subsets.}
  The top row compares ForeFly with existing methods on the Unseen Map subset, and the bottom row reports results on the Unseen Object subset.}
   \label{fig:unseen_map_object}
\end{figure}

As shown in Fig.~\ref{fig:unseen_map_object}, ForeFly consistently outperforms all comparison methods on both Unseen Map and Unseen Object across the evaluation metrics. On Unseen Map, the improvement is particularly pronounced in OSR, where ForeFly reaches 40.2\% compared with 30.3\% for LongFly, indicating stronger robustness to previously unseen environments. On Unseen Object, ForeFly also achieves consistent gains in navigation accuracy, success rate, and path efficiency. These results demonstrate that ForeFly generalizes effectively to both environmental and target-level distribution shifts.

\begin{figure}[t]
  \centering
  \includegraphics[width=\textwidth]{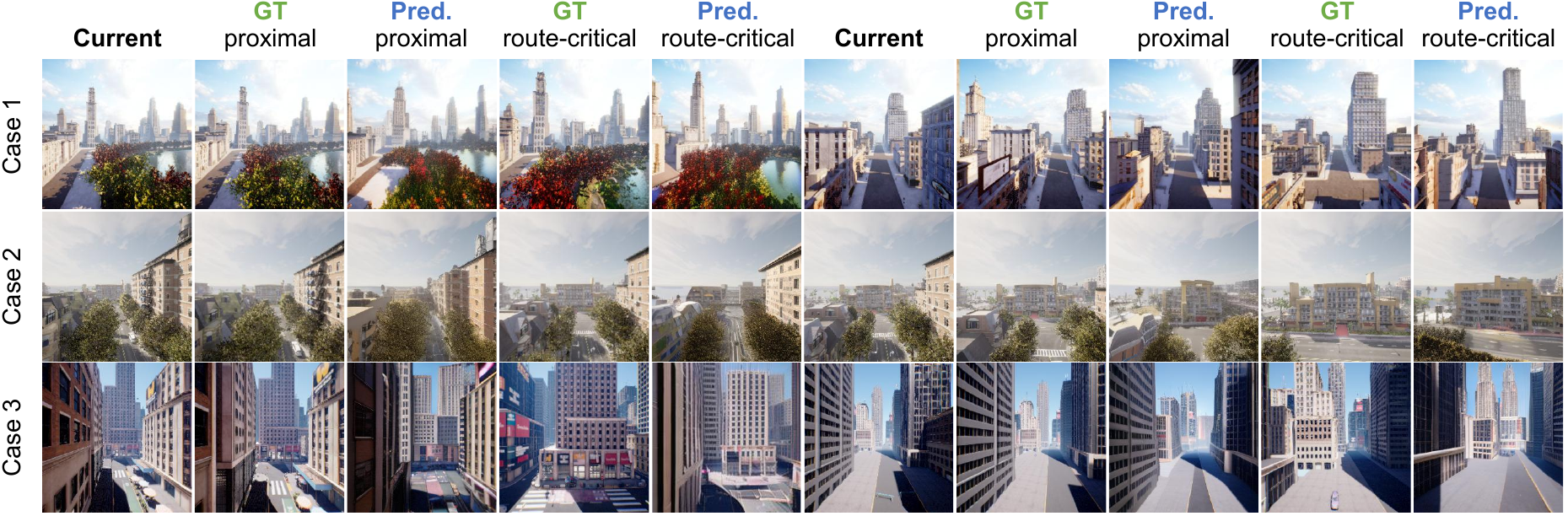}
  \caption{\textbf{Qualitative visualization of dual-horizon future prediction.}
    Each row shows two sampled waypoints from the same navigation trajectory. For each waypoint, we visualize the current observation together with the ground-truth and predicted proximal and route-critical futures.}
   \label{fig:vis_future}
\end{figure}

\subsection{Qualitative Analysis of Dual-Horizon Foresight}
\label{sec:qual_future}

To further examine whether the predicted dual-horizon latents capture meaningful future states, we provide an image-space visualization of ForeFly's foresight representations. Note that ForeFly itself predicts compact visual latents rather than RGB observations. For qualitative interpretation only, we employ an auxiliary latent-diffusion reconstructor that takes the predicted foresight representations as conditioning signals and reconstructs the corresponding future observations in image space. The reconstruction module is used solely for visualization and does not participate in navigation decision making.
Since the reconstruction is produced by an auxiliary generative model, its visual fidelity depends on both the quality of the predicted foresight representations and the capacity of the reconstructor. Therefore, these visualizations should be interpreted as qualitative evidence of the semantic and spatial information retained in the predicted representations, rather than as direct pixel-level future prediction.

As shown in Fig.~\ref{fig:vis_future}, each row presents two sampled waypoints
from the same navigation trajectory. At each waypoint, we visualize the current
observation together with the ground-truth and reconstructed proximal and
route-critical futures. The proximal predictions largely preserve the local
scene layout and viewpoint continuity, whereas the route-critical predictions
exhibit more pronounced changes in scene composition, reflecting farther
route progression.

Across all three cases, the reconstructed futures remain broadly consistent
with the corresponding ground-truth observations in their dominant spatial
and semantic structures. Moreover, the predictions evolve with the navigation
progress rather than collapsing to a fixed future pattern. These results
qualitatively demonstrate the complementary roles of the two foresight
horizons: proximal foresight captures short-term scene evolution, while
route-critical foresight anticipates longer-range navigation-relevant changes.

\section{Limitations and Future Directions}
\label{sec:supp_limitations}
ForeFly has several limitations that define directions for future work. First,
route-critical targets depend on turning and distance heuristics. The fixed
thresholds work well on TravelUAV, but different vehicle dynamics, sampling
rates, or environment scales may require recalibration or a learned event
detector. Second, each state is paired with one demonstrated future. The
squared latent objective does not explicitly represent multiple valid futures
when several routes satisfy an instruction. Probabilistic latent prediction
would be a natural extension. Third, retrospective memory is built from the
executed trajectory and can inherit errors after route deviation. Explicit
uncertainty estimates or memory recovery mechanisms could improve robustness
in such cases.

Real-world UAV experiments are important for validating practical applicability,
but large-scale aerial evaluation is constrained by safety, regulatory
compliance, privacy, and site-access requirements. Following the standard AVLN
protocol, we evaluate ForeFly on TravelUAV and UAV-ON, which provide diverse
large-scale outdoor environments, realistic visual observations, and seen/unseen
settings for controlled and reproducible evaluation. Existing methods such as
SimpleNav, NavFoM, AirForsight, LongFly, and SpatialFly are also primarily
evaluated in simulation, enabling fair comparison under a common protocol.
Future work will further investigate sim-to-real transfer and physical UAV
deployment to assess ForeFly under real-world conditions.

\end{document}